\documentclass[letterpaper, 10 pt, conference]{ieeeconf}  

\IEEEoverridecommandlockouts                              

\usepackage{graphics} 
\usepackage{epsfig} 
\usepackage{times} 
\usepackage{amsmath} 
\usepackage{amssymb}  
\usepackage{xcolor}
\definecolor{myyellow}{rgb}{1.0, 1.0, 0.8} 

\usepackage{algorithm}
\usepackage{algorithmic}
\usepackage{graphicx}
\usepackage{subcaption}
\usepackage{booktabs}
\usepackage{multirow}
\usepackage{makecell}
\usepackage{float}
\usepackage{url}

\newif\ifanonymous
\anonymousfalse     

\title{\LARGE \bf
Self-Curriculum Model-based Reinforcement Learning for Shape Control of Deformable Linear Objects
}

\ifanonymous
\author{Anonymous Authors%
\thanks{Affiliations and contact information are withheld for double-blind review.}%
}
\else
\author{Zhaowei Liang$^{1}$, Song Wang$^{1}$, Zhao Jin$^{1}$, Shirui Wu$^{1}$, and Dan Wu$^{1,*}$%
\thanks{$^{1}$All authors are with the Department of Mechanical Engineering, Tsinghua University, Beijing, China.}%
\thanks{*Corresponding author: Dan Wu, e-mail: wud@tsinghua.edu.cn}%
\thanks{This work was financially supported by the National Natural Science Foundation of China [Grant No.52375019].}%
}
\fi

\begin{document}
\bstctlcite{IEEEexample:BSTcontrol}

\maketitle
\thispagestyle{empty}
\pagestyle{empty}

\begin{abstract}

Precise shape control of Deformable Linear Objects (DLOs) is crucial in robotic applications such as industrial and medical fields. However, existing methods face challenges in handling complex large deformation tasks, especially those involving opposite curvatures, and lack efficiency and precision. To address this, we propose a two-stage framework combining Reinforcement Learning (RL) and online visual servoing. In the large-deformation stage, a model-based reinforcement learning approach using an ensemble of dynamics models is introduced to significantly improve sample efficiency. Additionally, we design a self-curriculum goal generation mechanism that dynamically selects intermediate-difficulty goals with high diversity through imagined evaluations, thereby optimizing the policy learning process. In the small-deformation stage, a Jacobian-based visual servo controller is deployed to ensure high-precision convergence. Simulation results show that the proposed method enables efficient policy learning and significantly outperforms mainstream baselines in shape control success rate and precision. Furthermore, the framework effectively transfers the policy trained in simulation to real-world tasks with zero-shot adaptation. It successfully completes all 30 cases with diverse initial and target shapes across DLOs of different sizes and materials.
\ifanonymous
The project website will be accessible upon acceptance.
\else
The project website is available at: \url{https://hunger-beat.github.io/sc-mbrl-dlo/}
\fi
\end{abstract}

\section{INTRODUCTION}

Deformable linear objects (DLOs), such as wires, cables, and sutures, are widely used in electronic assembly, medical procedures, and industrial automation. Precise shape control of DLOs is a common and essential manipulation task in these scenarios. Automating this process with robots presents significant challenges due to the high degrees of freedom, nonlinear dynamics, and complex material properties of DLOs~\cite{gu2025learning,wang2025self,wu2024optimization}.

\begin{figure}[t]
    \centering
    \includegraphics[width=0.90\columnwidth]{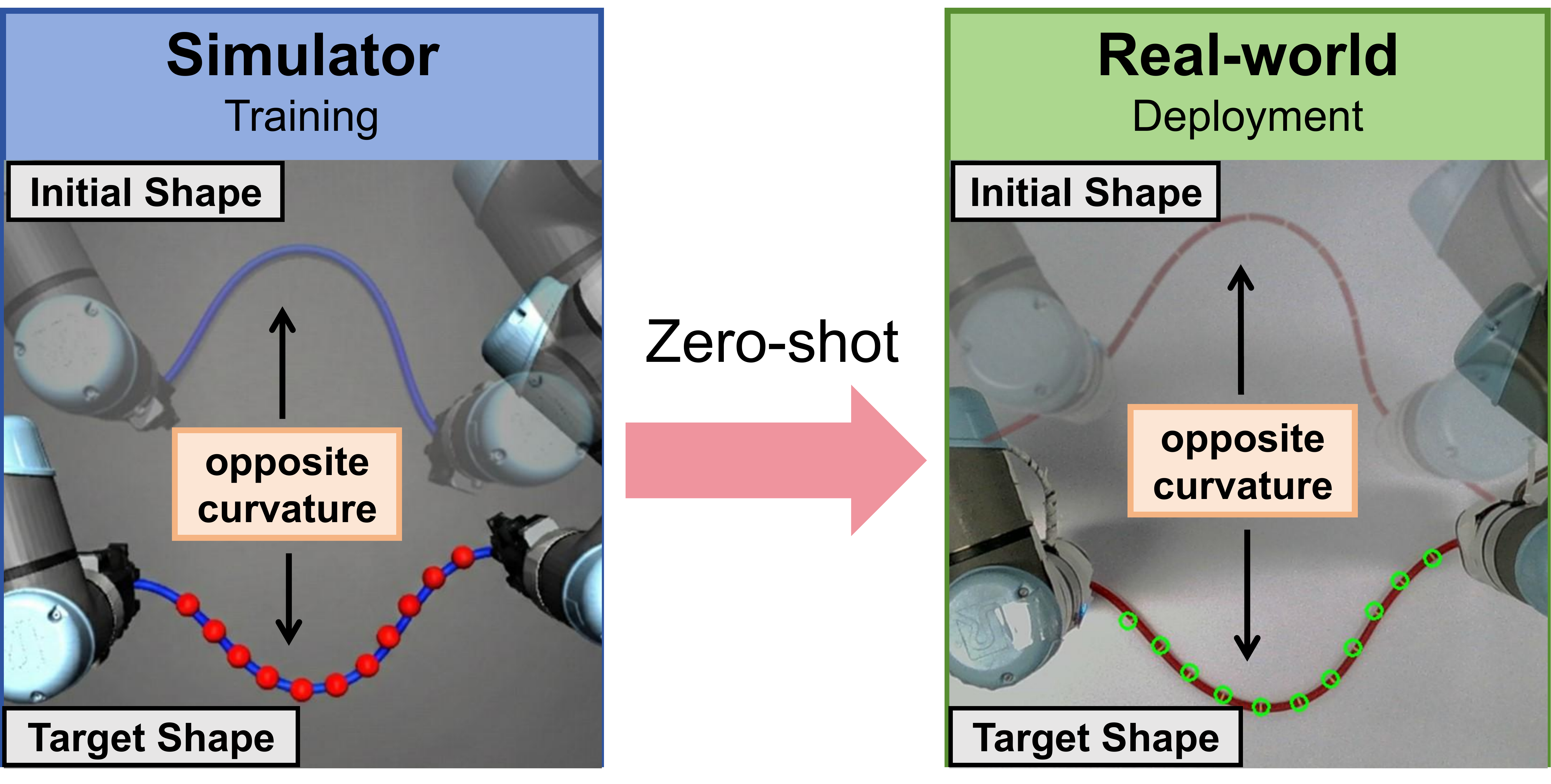}  
    \caption{Illustration of the task with large deformations exhibiting opposite curvatures. The initial and target shapes have completely opposite concavities. The policy is trained in a simulator and transferred to the real-world scenario.}
    \label{fig:fig1}
\end{figure}

In DLO shape control research, early efforts primarily focused on small-deformation scenarios~\cite{zhu2018dual,lagneau2020automatic}. Subsequently, studies gradually expanded to large‑deformation shape control~\cite{yu2022global,almaghout2024robotic,daniel2023multi}, achieving notable progress. However, when the initial and target shapes differ substantially—especially in cases of completely opposite curvature (see Fig.~\ref{fig:fig1})—the task difficulty increases significantly: they require long-horizon planning and careful handling to avoid overstretching near straight configurations~\cite{yu2022global}. Nevertheless, very few studies have focused on these complex deformations, and existing methods also have limitations. Almaghout et al.~\cite{almaghout2024robotic} tackles large deformations by planning a sequence of intermediate shapes, yet the control process takes several minutes; Laezza et al.~\cite{laezza2024offline} adopts an offline‑learning policy but relies on extensive real‑world interaction data and ultimately achieves limited shape‑control accuracy. These shortcomings highlight a fundamental gap in efficiently and precisely handling complex large-deformation tasks, especially under the challenging opposite-curvature condition.

Reinforcement learning (RL) has demonstrated potential in deformable object manipulation~\cite{yin2021modeling}, but its application to DLO shape control still faces multiple challenges. First, the inherently low sample efficiency of RL severely limits its applicability: existing methods either require excessively long training times~\cite{daniel2023multi} or depend on expensive real‑data collection~\cite{laezza2024offline}. Second, to achieve flexible shape control, one often aims for a single policy that can adapt to diverse target shapes, framing the problem as Goal-Conditioned Reinforcement Learning (GCRL)~\cite{liu2022goal}, where the goal is the desired DLO shape. However, this formulation worsens the sample-efficiency issue. Moreover, most RL approaches remain confined to simulation~\cite{laezza2021learning,zakaria2022robotic}, with significant difficulties in sim‑to‑real transfer and cross-DLO generalization. These challenges hinder the application of RL to real-world DLO manipulation, especially in high-precision tasks under large deformations.

To address the challenges mentioned above, this paper proposes a two‑stage shape control framework that integrates model‑based RL with visual servoing, as shown in Fig~\ref{fig:fig2}. In the large-deformation stage, an RL policy is trained in a simulator, utilizing an ensemble of dynamics models to improve RL sample efficiency. A self-curriculum goal generation mechanism, which balances goal difficulty and diversity, provides continuous moderate interaction goals for the policy, enabling the efficient training of a general policy capable of handling a variety of initial and target shapes. In the small-deformation stage, an online Jacobian-based visual servoing controller further refines the control process, ensuring precise convergence. The policy trained purely in simulation can be transferred directly to the real world, and is validated on various DLOs. The contributions of this article can be summarized as follows:

\begin{itemize}
    \item We propose a novel two-stage shape control framework capable of handling large deformations, including challenging opposite-curvature cases, with high efficiency and precision.
    \item We introduce a model‑based RL framework that significantly improves sample efficiency. We further design a self‑curriculum goal generation method, enabling the policy to handle varied initial and target shapes.
    \item We validate our method extensively in simulation and demonstrate zero-shot transfer to real-world experiments, achieving robust shape control across DLOs of different materials and sizes, which shows strong generalization.
\end{itemize}

\section{RELATED WORK}

\subsection{DLO Shape Control}

Existing DLO shape control methods fall into physics-based and data-driven categories~\cite{yu2022global}. Physics-based methods struggle with accurate parameter estimation in real settings~\cite{wu2024optimization}. Consequently, data-driven methods have gained increasing attention.

A prominent approach uses Model Predictive Control (MPC), which typically learns a DLO dynamics model and solves via online optimization. For instance, Gu et al.~\cite{gu2025learning} proposed GA-Net to learn the DLO dynamics, enabling control of unseen DLOs. To handle shape control with external contacts, Huang et al.~\cite{huang2023learning} introduced an action generation module combining random sampling and gradient descent. While effective, MPC-based approaches suffer from computationally expensive online optimization, leading to low efficiency in dynamic environments~\cite{shi2024robocraft}. To address this, we learn a policy guided by a dynamics model.

RL methods learn direct mappings from states to actions. Laezza et al.~\cite{laezza2021learning} designed a novel reward function and used DDPG to control elastoplastic DLOs in simulation; Daniel et al.~\cite{daniel2023multi} extended manipulation to large-strain DLOs by decomposing the action space for single-arm 3D deformation. Beyond the sample inefficiency and simulation confinement mentioned above, standalone RL methods for large-deformation shape control also exhibit limited final precision~\cite{laezza2024offline}, making consistently high-accuracy convergence challenging.

Another major category is Jacobian-based methods, which compute robot motions using simplified DLO deformation models. Some works update the Jacobian model numerically, e.g., via least‑squares estimation~\cite{zhu2018dual,jin2019robust}, while others derive it analytically~\cite{almaghout2024robotic}. These methods require no prior training and generalize well, but they are effective only for small deformations and are prone to local optima in large‑deformation tasks.

Our approach combines these strengths: RL handles large and complex deformations while Jacobian-based visual servoing ensures precision in small-deformation regimes. We also further introduce a model-based RL (MBRL) framework to tackle the critical sample inefficiency of RL. Unlike~\cite{yang2022online}, which uses MBRL but relies on inefficient MPC for online planning, our method learns the policy directly.

\subsection{Self-Curriculum Goal-Conditioned Reinforcement Learning}

Self-curriculum learning, which allows the agent to learn on a sequence of goals  arranged from easy to difficult, is a promising approach to the sample efficiency issue in GCRL~\cite{gong2024goal}. Existing methods can be broadly categorized into novelty-based and difficulty-based strategies~\cite{castanet2023stein}.

Novelty-based goal generation aims to select goals based on state visit density, with the objective of expanding the set of achieved goals~\cite{kim2023variational,gong2024goal}. However, these approaches lack an explicit evaluation of the policy's current capability, ignoring goal reachability~\cite{castanet2023stein}. Given these limitations, we thus focus on the difficulty-based paradigm, offering a more structured approach.

Difficulty-based methods assume that optimal learning efficiency is achieved when goals are slightly beyond the agent's current skill level, termed intermediate-difficulty goals~\cite{liu2022goal}. A common approach employs adversarial learning to generate such goals dynamically. For example, Goal GAN~\cite{florensa2018automatic} uses a generator-discriminator framework to produce goals matching the agent's capability. However, adversarial training often suffers from instability and imbalance. To mitigate this, Castanet et al.~\cite{castanet2023stein}trains a supervised reachability predictor to generate intermediate-difficulty goals, though their generator struggles with high-dimensional goal spaces. Other methods, such as~\cite{kim2023variational}, leverage epistemic uncertainty from ensemble Q-functions to identify intermediate-difficulty goals.

Nevertheless, despite the progress made by these methods, many approaches condition goal generation on a fixed initial state and do not validate their methods in environments with diverse initial states~\cite{kim2023variational}. In contrast, we propose a model-based approach that periodically evaluates the policy using a learned dynamics model and samples goals that are both moderately difficult and diverse according to the current policy's capability. This method can handle diverse initial state scenarios, making it more suitable for our task.

\section{PROBLEM STATEMENT}

This paper addresses the shape control problem of elastic DLOs. Two robotic arms are fixed at the two ends of the DLO and collaboratively manipulate the object in a 2D plane, transforming it from an initial shape to a target shape. The shape of the DLO is represented by a set of uniformly distributed key feature points along its length. The task is defined as controlling the position of the end-effectors to move the feature points on the DLO to the corresponding desired positions. In particular, we aim to solve the problem of complex large deformations, especially those in which the initial and target shapes exhibit completely opposite curvatures, as shown in Fig.\ref{fig:fig1}.

We make the following assumptions to narrow the scope of the problem:
\begin{enumerate}
    \item Each robot's motion is constrained to 3 degrees of freedom, namely translation along the x and y axes, and rotation around the z-axis, giving a total of 6 degrees of freedom for the two robots.
    \item A vision-based algorithm is available to track the feature points in real-time with high accuracy.
    \item Similar to \cite{laezza2024offline}, the end-effectors of the two robots are confined within separate safe bounds to prevent collisions. A task fails if an end-effector goes out of bounds or if the distance between the end-effectors becomes too large, causing the DLO to be overstreched.
    \item The DLOs considered in this study undergo primarily elastic deformations, and the approach is not applicable to DLOs with extremely low stiffness.

\end{enumerate}

The notation for the main variables is as follows. The $i$-th keypoint is denoted as $\mathbf{x}_i \in \mathbb{R}^2$, and the shape of the DLO is represented as $\mathbf{X} := [\mathbf{x}_1, \dots, \mathbf{x}_N] \in \mathbb{R}^{2N}$, where $N$ is the number of feature points. The target shape is denoted as $\mathbf{X}^d := [\mathbf{x}_1^d, \dots, \mathbf{x}_N^d] \in \mathbb{R}^{2N}$, where $\mathbf{x}_i^d$ represents the desired position of the $i$-th keypoint. The pose of the end-effectors is defined by $\mathbf{r} = [\mathbf{r}_1, \mathbf{r}_2] \in \mathbb{R}^6$, where $\mathbf{r}_m = [r_{mx}, r_{my}, \phi_{m}]$, and $m = 1$ or $2$. It is important to note that the target shape does not specify the desired robot pose.

\section{Method}

\subsection{Two-Stage Shape Control Framework}

\begin{figure}[t]
    \centering
    \includegraphics[width=0.90\columnwidth]{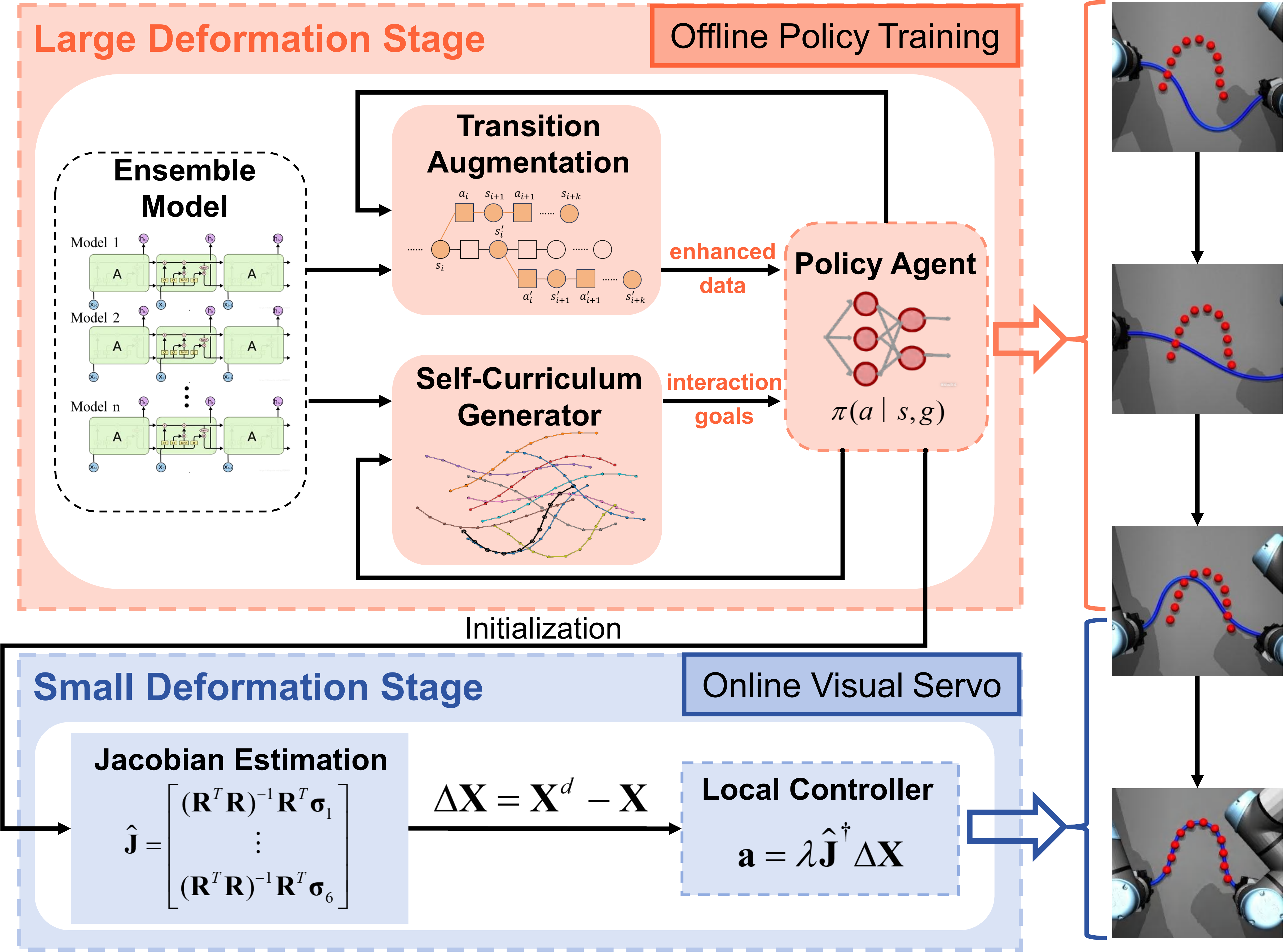}  
    \caption{Overview of the proposed DLO shape control framework. In the large-deformation stage, policy is trained with transition augmentation using ensemble model, while a self-curriculum module adaptively selects interaction goals. In the small-deformation stage, an online visual servo controller refines the motion to achieve precise convergence.}
    \label{fig:fig2}
\end{figure}

To achieve efficient and precise shape control for DLOs under large deformations, we propose a two-stage control framework, as illustrated in Fig.~\ref{fig:fig2}. The task is divided into a large-deformation stage and a small-deformation stage based on the shape control error e, defined as the root mean square error (RMSE) between corresponding feature points:
\begin{equation}
    e = \sqrt{\frac{1}{N} \sum_{i=1}^{N} | \mathbf{x}_i - \mathbf{x}_i^d |_2^2}
\end{equation}
The stage transition is triggered when $e$ falls below a predefined threshold $\epsilon$. In the large-deformation stage, an RL agent is trained in the simulator. Once $e < \epsilon$, the system switches to the small-deformation stage, where a Jacobian-based visual servo controller is applied online to minimize the shape error.

In the small-deformation stage, we adopt the method from \cite{zhu2018dual} for the online estimation of the Jacobian matrix ${\mathbf{J}}$. This matrix is iteratively updated using a weighted least-squares approach \cite{jin2019robust}. The action is then computed as:
\begin{equation}
\Delta \mathbf{r} = \lambda \hat{\mathbf{J}}^{\dagger} (\mathbf{X}^d - \mathbf{X})
\end{equation}
where $\hat{\mathbf{J}}^{\dagger}$ denotes the pseudo-inverse of the estimated Jacobian matrix and $\lambda$ is the gain parameter.

Notably, during the execution of the RL policy in the large-deformation stage, the Jacobian matrix is continuously updated. This provides a well-initialized estimate for the subsequent small-deformation phase, which mitigates the sensitivity of the visual servo method to random initial estimates and facilitates precise convergence.

\subsection{Model-Based Policy Learning}

\begin{figure}[t]
    \centering
    \includegraphics[width=0.90\columnwidth]{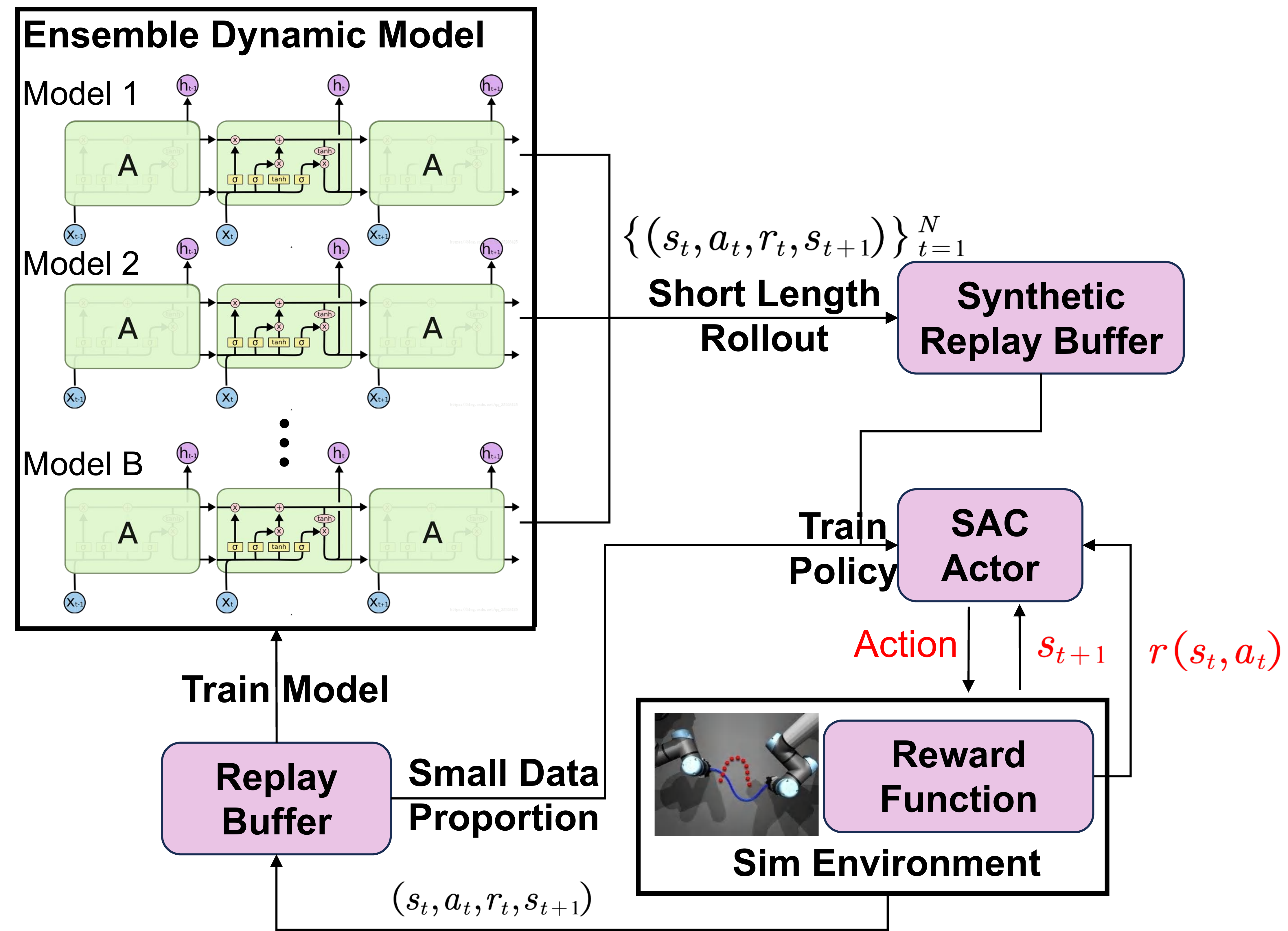}  
    \caption{The architecture of the MBRL framework with ensemble dynamics models.}
    \label{fig:fig3}
\end{figure}

In the large-deformation stage, to achieve sample-efficient RL policy training, we build upon the Model-based Policy Optimization (MBPO) framework \cite{janner2019trust} and extend it to the DLO manipulation task. The training architecture is illustrated in Fig.~\ref{fig:fig3}. Each  epoch, we train an ensemble of dynamics models using data from environment interactions. The trained models are then used to generate synthetic data, augmenting the policy training set and thereby improving overall data efficiency. Before training starts, we populate the replay buffer by collecting a small initial dataset using a random policy.

To effectively capture the dynamics of robot–DLO interaction, we employ a bi‑directional LSTM (Bi‑LSTM) network \cite{yan2020self} to model the sequential, linear structure of the DLO. The model maps the current DLO shape $\mathbf{X_t}$ and the robot action $\Delta \mathbf{r_t}$ to the parameters of a Gaussian distribution that predicts the state increment:
\begin{equation}
    [\boldsymbol{\mu}_\Delta, \boldsymbol{\sigma}_\Delta] = f_\theta(\mathbf{X}_t, \Delta \mathbf{r}_t)
\end{equation}
where $\boldsymbol{\mu}_\Delta$ and $\boldsymbol{\sigma}_\Delta$ are the mean and standard deviation of the predicted shape change $\Delta\mathbf{X}_{t} = \mathbf{X}_{t+1} - \mathbf{X}_{t}$. This defines the conditional probability density $p(\mathbf{X}_{t+1} \mid \mathbf{X}_t, \Delta \mathbf{r}_t)$ of the next state $\mathbf{X_{t+1}}$. We construct an ensemble prediction model consisting of B such sub-models, where B is the ensemble size, each independently learning the dynamics mapping. The loss function for each probabilistic model is derived from Maximum Likelihood Estimation (MLE).

After each training round, we select the top B' sub-models with the smallest validation losses from the ensemble to form an elite set, where B' is the number of elite models, enhancing prediction accuracy. Subsequently, a short branched rollout is performed using these elite models: initial states are sampled from the real experience replay buffer; actions are generated by the current policy; and states are predicted by dynamically selecting models from the elite set. This process generates batches of synthetic training data in parallel. Finally, real and synthetic data are mixed to train the policy using the Soft Actor-Critic (SAC) algorithm in an off-policy manner, enabling both sample-efficient and stable policy learning.

In this task, the action space $\mathcal{A}$ of the policy is defined as the displacement increment $\Delta \mathbf{r}$ of the robotic end-effectors. The observation space $\mathcal{O}$ is formed by concatenating the current DLO shape $\mathbf{X}$, the current end-effector poses $\mathbf{r}$, and the target shape $\mathbf{X}^d$. A task is considered successful if the shape control error $e$ falls below a stricter threshold $\epsilon_{\text{RL}}$ (with $\epsilon_{\text{RL}} < \epsilon$). The reward function is a dense reward consisting of: 1) a negative reward proportional to $e$; 2) incremental positive rewards when $e$ decreases below multiple predefined thresholds; and 3) a significantly large positive reward upon task success.

\subsection{Self-Curriculum Goal Generation}

\begin{figure}[t!]
    \centering
    
    \begin{subfigure}[b]{0.354\columnwidth}
        \centering
        \includegraphics[height=0.1\textheight, keepaspectratio]{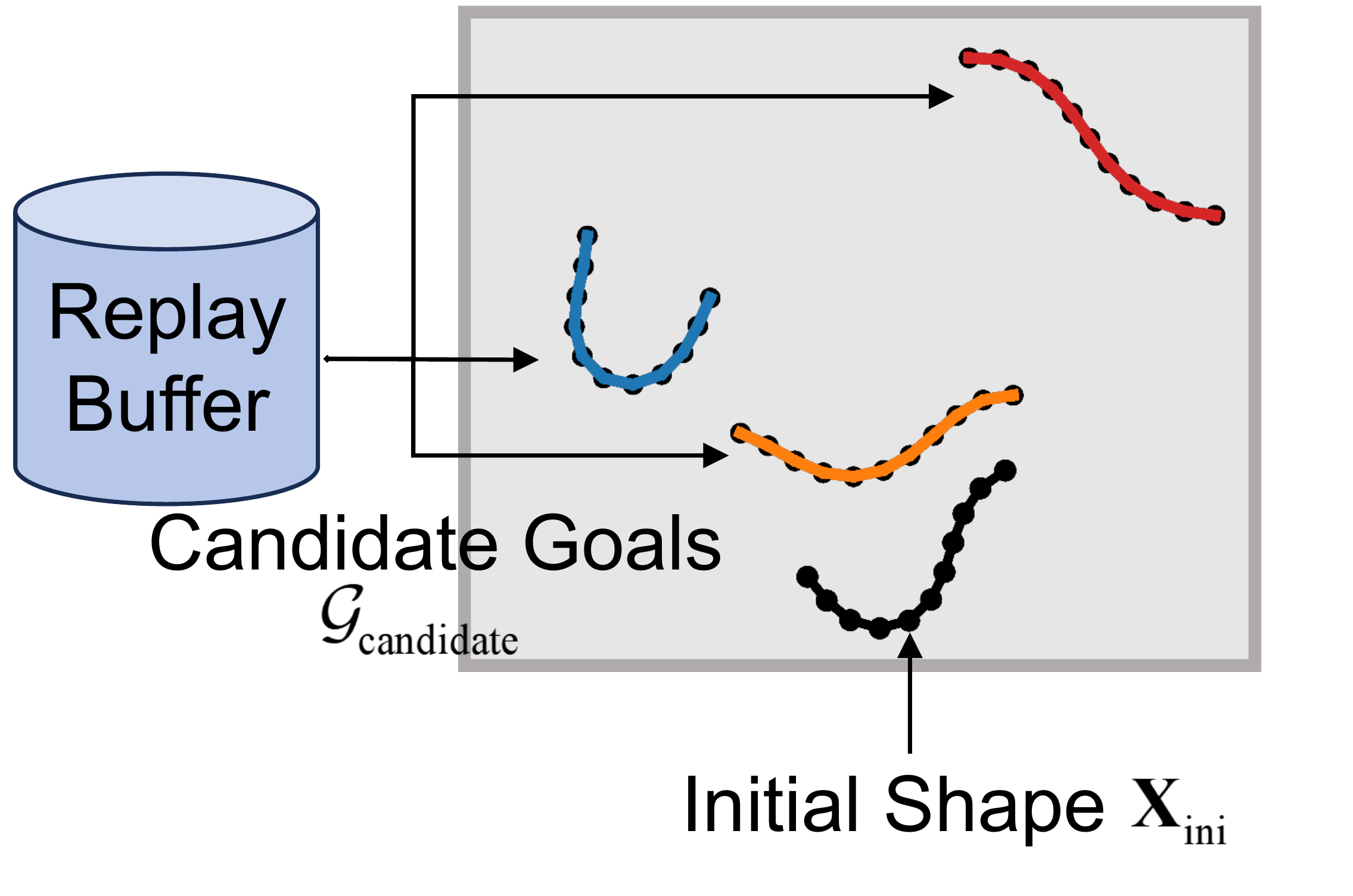}
        \caption{}
        \label{fig:pic4a}
    \end{subfigure}
    \hfill
    \begin{subfigure}[b]{0.555\columnwidth}
        \centering
        \includegraphics[height=0.1\textheight, keepaspectratio]{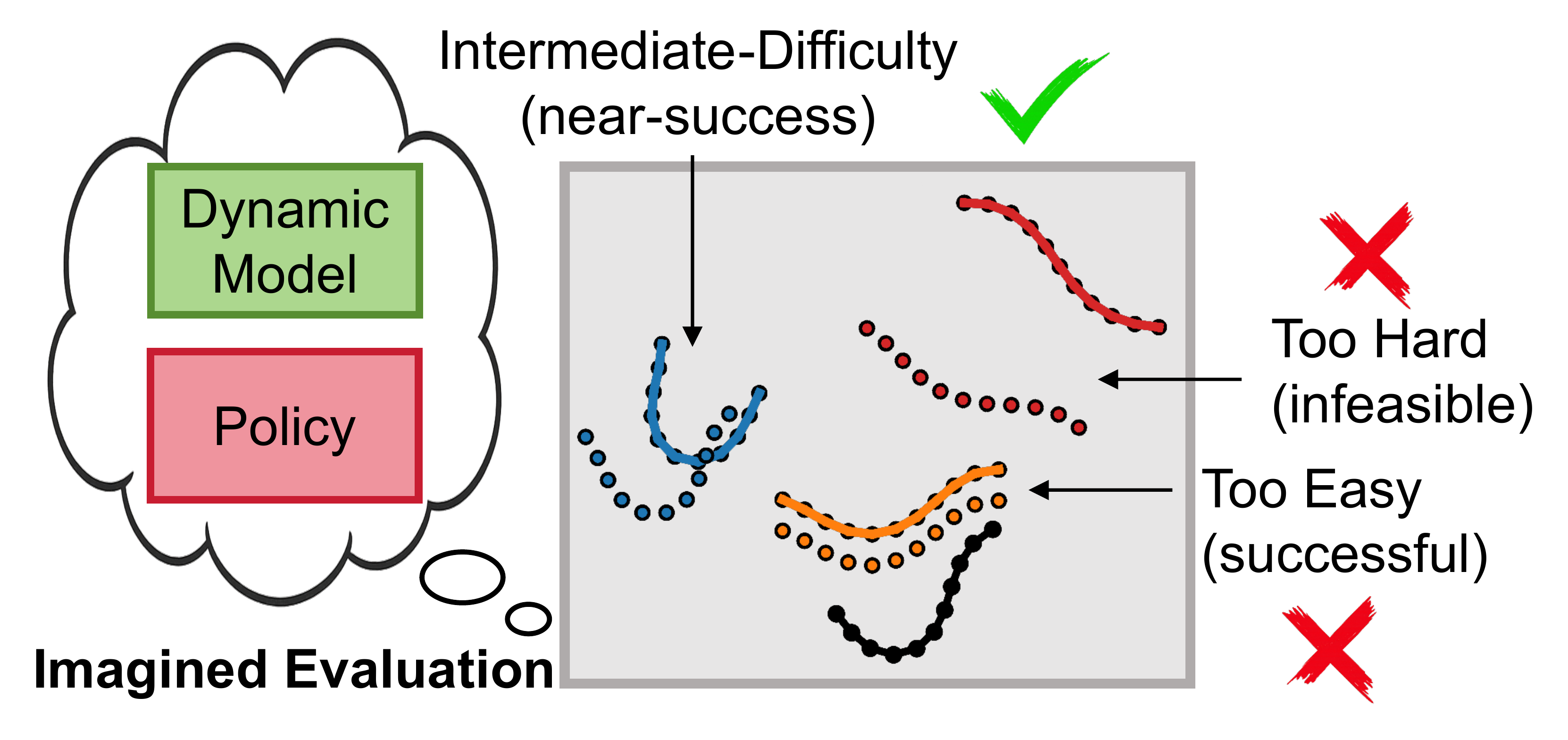}
        \caption{}
        \label{fig:pic4b}
    \end{subfigure}
    
    \vspace{0.2cm} 
    
    \begin{subfigure}[b]{0.91\columnwidth}
        \centering
        \includegraphics[width=\linewidth]{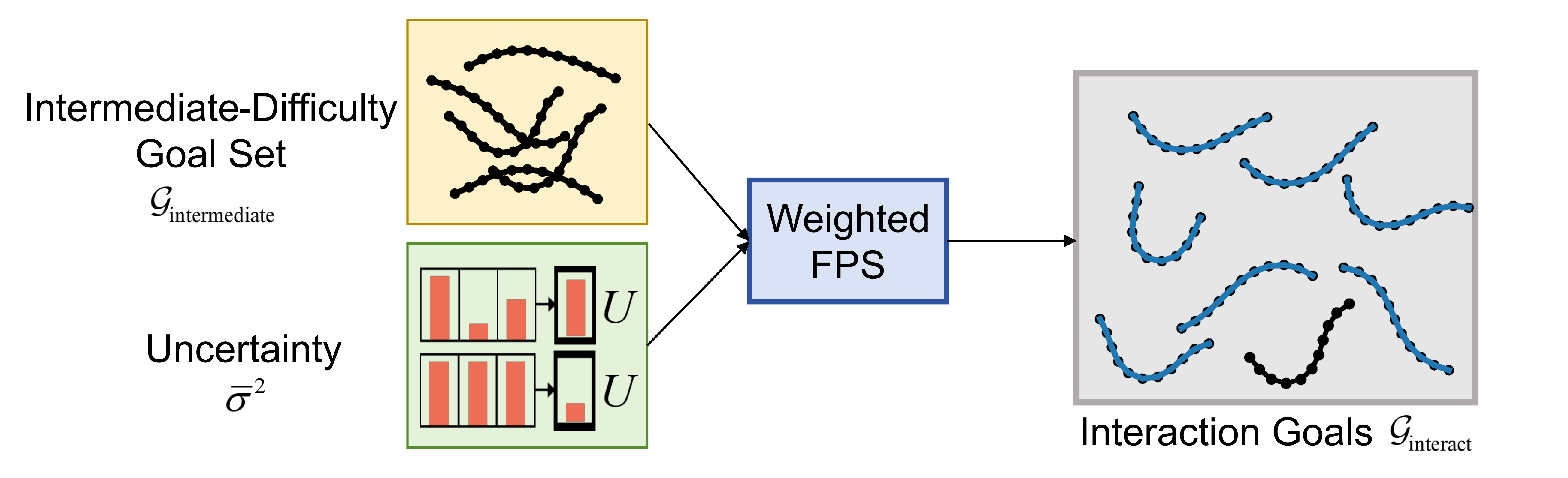}
        \caption{}
        \label{fig:pic4c}
    \end{subfigure}
    
    \caption{The proposed self-curriculum goal generation method. (a) Candidate goals are randomly sampled from the replay buffer. (b) Imagined evaluations are conducted to categorize intermediate-difficulty goals. (c) Weighted Farthest Point Sampling (FPS) selects the final interaction goal set $\mathcal{G}_{\text{interact}}$.}
    \label{fig:fig4}
\end{figure}

At the beginning of each training epoch, goals for the environment interactions in that epoch are selected. We posit that stable and efficient shape control learning can only be achieved by continuously generating goals that are of moderate difficulty, highly diverse, and informative. To this end, we propose a self-curriculum goal generation method based on model-based imagined evaluation, which adaptively constructs the goal set for environment interaction in each epoch, as illustrated in Fig.~\ref{fig:fig4}.

In this process, we first randomly sample M historically reached DLO states from the replay buffer to form a candidate goal set, denoted as $\mathcal{G}_{\text{candidate}}$, as shown in Fig.~\ref{fig:pic4a}. For each candidate goal $\mathbf{X}^d$, the algorithm conducts imagined look-ahead evaluations using the current policy and the current ensemble dynamics model. During evaluation, the policy, conditioned on the candidate goal $\mathbf{X}^d$, generates action sequences starting from the current epoch's initial state $\mathbf{X}_\text{ini}$. To enhance prediction robustness, each state transition is given by the mean prediction of all elite models. For each candidate goal, we perform K independent evaluation trajectories, record the minimum error $e_{\text{min}}^{(k)}$ between the DLO shape and the target shape during each trajectory, and compute the average as the candidate's mean minimum error $\bar{e}_{\text{min}}$.

We define an intermediate-difficulty goal as one whose mean minimum error $\bar{e}_{\text{min}}$ is slightly above the task success threshold $\epsilon_{\text{RL}}$. In other words, such a goal is challenging yet within reach for the current policy, thereby providing the most informative training signal. The set of intermediate-difficulty goals meeting this criterion is formally defined as:

\begin{equation}
    \mathcal{G}_{\text{intermediate}} = \left\{ \mathbf{X}^d  \;\middle|\; \epsilon_{\text{RL}} < \bar{e}_{\text{min}}(\mathbf{X}^d) < \epsilon_{\text{upper}} \right\}
\end{equation}
where $\epsilon_{\text{upper}}$ is a predefined upper error bound. The imagined evaluation process is depicted in Fig.~\ref{fig:pic4b}.

Concurrently, during the imagined evaluation, we compute the variance $\sigma^2$ among the predictions of different elite models at each step. For a candidate goal, the average prediction variance $\bar{\sigma}^2$ across its evaluation trajectories reflects the model's epistemic uncertainty in the surrounding state space. A higher $\bar{\sigma}^2$ indicates a less learned region, and selecting such goals for training is expected to maximally improve the model's generalization.

After identifying the intermediate-difficulty
 goal set $\mathcal{G}_{\text{intermediate}}$ and its corresponding variances $\bar{\sigma}_i^2$, we proceed to select a final subset of goals for environment interaction. To avoid an uneven distribution that harms policy generalization, we employ a Weighted Farthest Point Sampling (FPS) strategy (Algorithm~\ref{alg:1}). This strategy ensures that the selected goals are well-distributed in the shape space while incorporating prediction uncertainty as sampling weights. On one hand, the FPS mechanism encourages selecting goals that are far apart from each other, promoting diversity. On the other hand, the uncertainty weight biases the sampling towards goals with higher prediction variance, guiding exploration towards poorly modeled regions.

Through the above self-curriculum goal generation process, we obtain the final goal set $\mathcal{G}_{\text{interact}}$ for the current epoch. In the subsequent environment interaction phase of this epoch, interaction goals are randomly sampled from this set, enabling efficient and sustained performance improvement.

\begin{algorithm}[t]
\caption{Weighted Farthest Point Sampling}
\label{alg:1}
\begin{algorithmic}[1]

\REQUIRE Intermediate-difficulty goal set $\mathcal{G}_{\text{intermediate}}$, uncertainties $\{\bar{\sigma}_i^2\}$, number of goals to select $N_g$, weight $\alpha \in [0,1]$
\ENSURE Selected goal set $\mathcal{G}_{\text{interact}}$

\STATE Normalize uncertainties to obtain weights $\{w_i\}$
\STATE Randomly select the first goal $j_1$, initialize selected set $\mathcal{I}_{\text{selected}} = \{j_1\}$
\STATE Initialize minimum distances $d_{\min}(i) = \infty$ for all goals

\FOR{$k = 2$ to $N_g$}
    \STATE Update $d_{\min}(i) = \min(d_{\min}(i), \left\|\mathbf{X}_i^d - \mathbf{X}_{j_{k-1}}^d\right\|_2)$
    \STATE Compute scores: $s_i = \alpha \cdot \frac{d_{\min}(i)}{\max_j d_{\min}(j)} + (1-\alpha) \cdot w_i$
    \STATE Select $j_k = \arg\max_{i \notin \mathcal{I}_{\text{selected}}} s_i$
    \STATE $\mathcal{I}_{\text{selected}} \leftarrow \mathcal{I}_{\text{selected}} \cup \{j_k\}$
\ENDFOR

\STATE $\mathcal{G}_{\text{interact}} = \{\mathbf{X}_i^d \mid i \in \mathcal{I}_{\text{selected}}\}$
\RETURN $\mathcal{G}_{\text{interact}}$

\end{algorithmic}
\end{algorithm}

\section{SIMULATION AND RESULTS}

\subsection{Simulation Setup}

We developed a MuJoCo-based simulation environment for training and evaluating DLO shape control. The setup consists of two UR5e robotic arms, each equipped with a gripper. The DLO is modeled as a chain of 40 articulated capsules with a total length of 0.5~m, a diameter of 10~mm, a bending stiffness of \(5 \times 10^6\), and a damping coefficient of 0.2. A set of 13 key points is defined along the DLO, including both endpoints that are grasped by the grippers. Their positions are obtained directly from the simulator.

\subsection{Policy Learning Results}

To evaluate the effectiveness of our method, we train RL policies under two distinct initial conditions. The first condition starts the DLO from a straight-line configuration at the workspace center. The second condition begins with the DLO in varied initial shapes, randomly sampled from a pre-collected dataset of diverse configurations. In both cases, the agent autonomously generates interaction goals during training. For a fair evaluation of policy performance, we construct a separate fixed test set of 50 random target shapes for each condition and periodically evaluate the success rate throughout training.

The RL success threshold $\epsilon_{\text{RL}}$ is set to 20~mm. Key algorithm parameters are set as follows: the ensemble uses B=7 models with B'=5 elites; the Bi-LSTM has one layer of 256 hidden units; policy training samples data with a 99\%–1\% split between the synthetic and real buffers per epoch (3000 steps). For self-curriculum generation, we use M=5000, K=5, $\epsilon_{\text{upper}}=30~\text{mm}$, $\alpha=0.8$, and $N_g=20$. All experiments were conducted on a workstation with an Intel Core i7-13700 CPU and an NVIDIA GeForce RTX 4080 GPU.

\begin{figure}[t!]
    \centering
    \begin{subfigure}[b]{0.45\columnwidth}
        \centering
        \includegraphics[width=\linewidth]{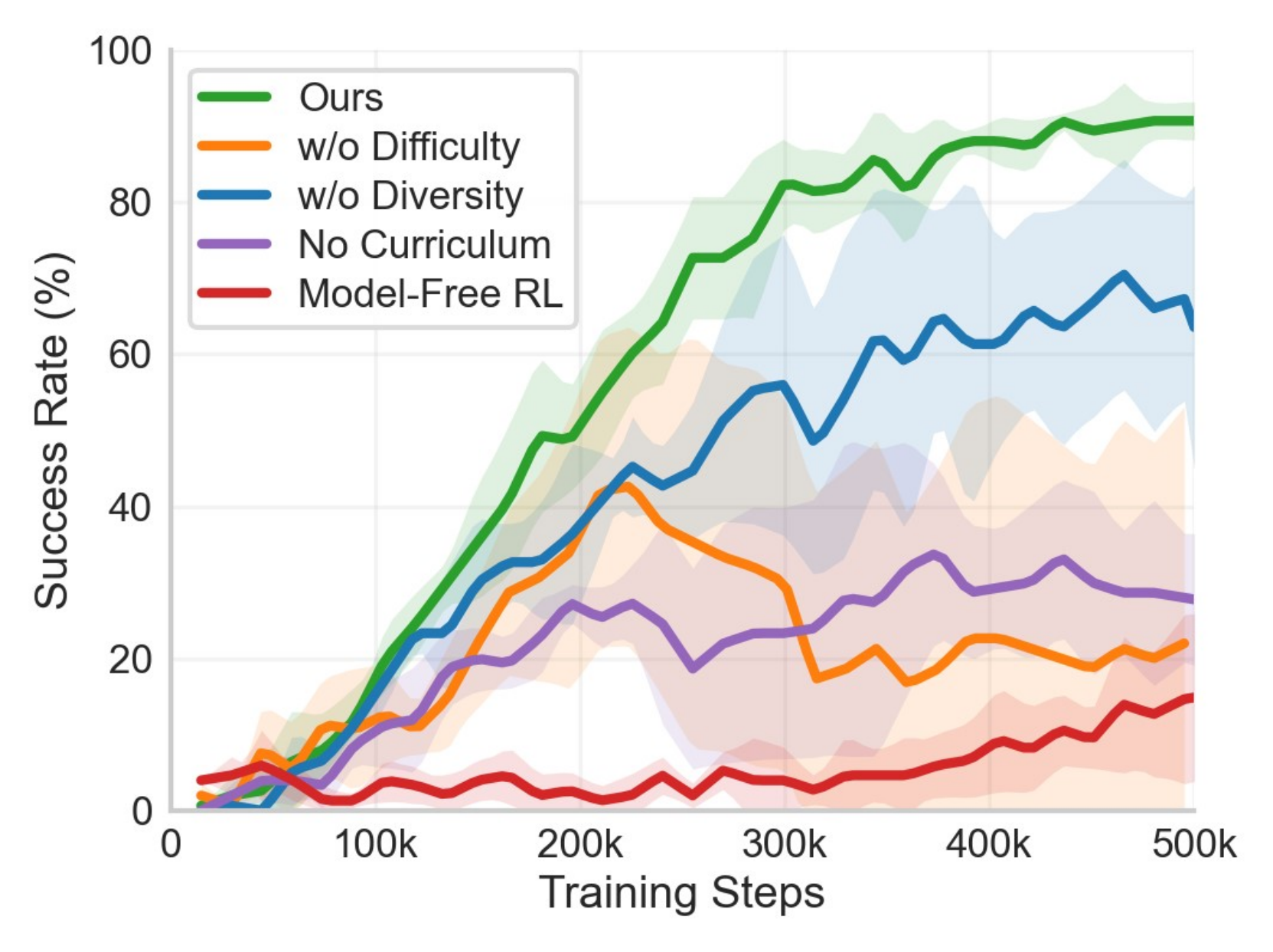}
        \caption{Straight Initial}
        \label{fig:pic5a}
    \end{subfigure}
    \hfill
    \begin{subfigure}[b]{0.45\columnwidth}
        \centering
        \includegraphics[width=\linewidth]{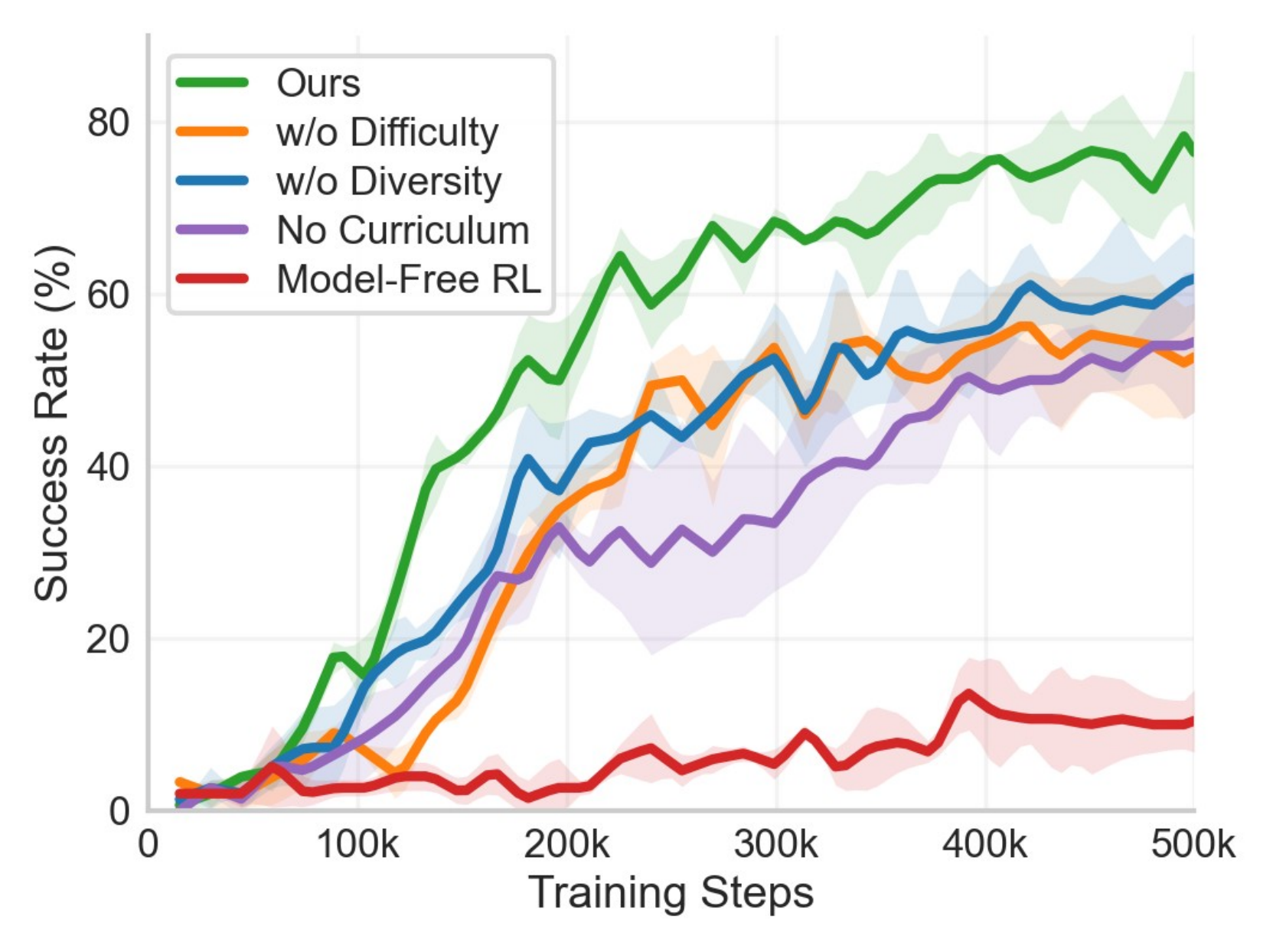}
        \caption{Diverse Initial}
        \label{fig:pic5b}
    \end{subfigure}

    \caption{Policy learning curves under two initial conditions: (a) straight-line initialization and (b) diverse-shape initialization.}
    \label{fig:fig5}
\end{figure}

We conduct ablation studies and baseline comparisons to validate our RL approach. First, we ablate key components of our self-curriculum scheme: (i) the difficulty filtering module and (ii) the diversity selection module (Weighted FPS). A variant without any curriculum learning, where goals are randomly sampled from replay buffer, is also included. Second, to demonstrate MBRL's sample efficiency, we compare against a model-free baseline trained solely on real replay buffer transitions. All results are averaged over three random seeds, and the learning curves are reported in Fig.~\ref{fig:fig5}.

\begin{figure}[t]
    \centering
    \begin{subfigure}[b]{0.90\columnwidth}
        \centering
        \includegraphics[width=\linewidth]{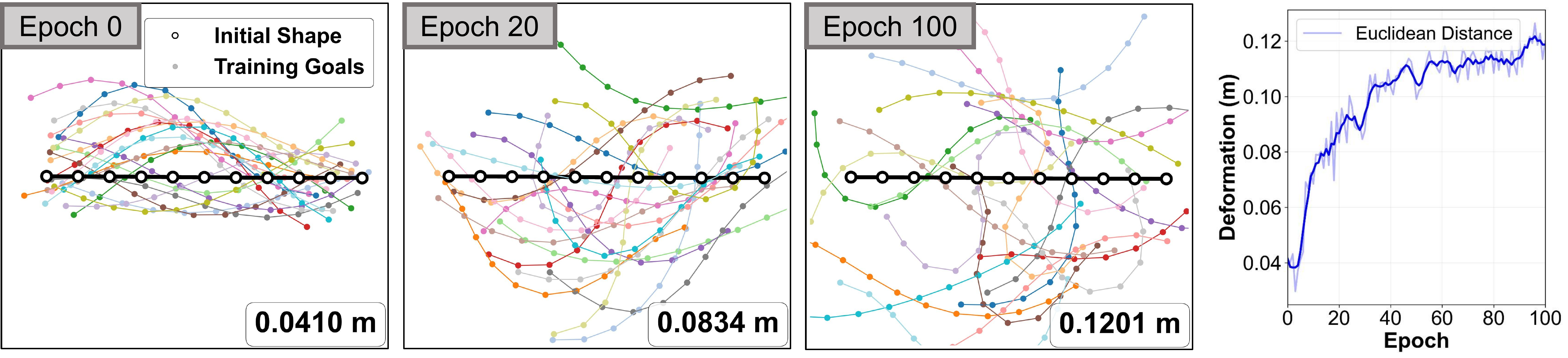}
        \caption{Straight Initial}
        \label{fig:pic6a}
    \end{subfigure}

    \vspace{1mm}

    \begin{subfigure}[b]{0.90\columnwidth}
        \centering
        \includegraphics[width=\linewidth]{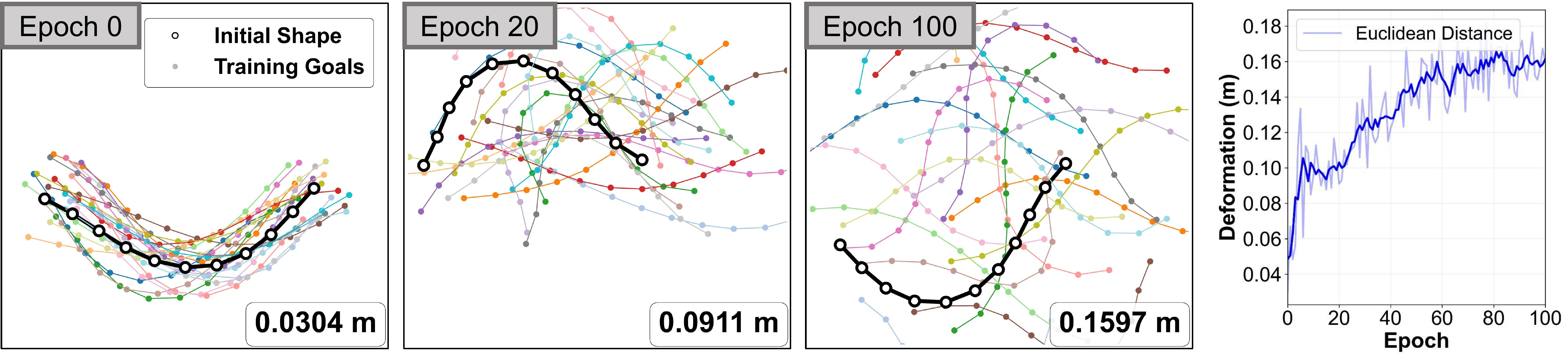}
        \caption{Diverse Initial}
        \label{fig:pic6b}
    \end{subfigure}

    \caption{Self-curriculum goal evolution under two initial conditions; the rightmost plot shows the average goal deformation versus epoch.} 
    \label{fig:fig6}
\end{figure}

\begin{figure*}[t!]   
  \centering
  \begin{minipage}[c]{\textwidth}
    \centering
    \makebox[0pt][r]{\raisebox{0.5\height}{(a)}\hspace{6pt}}%
    \begin{subfigure}[c]{0.92\textwidth}
      \centering
      \includegraphics[width=\linewidth]{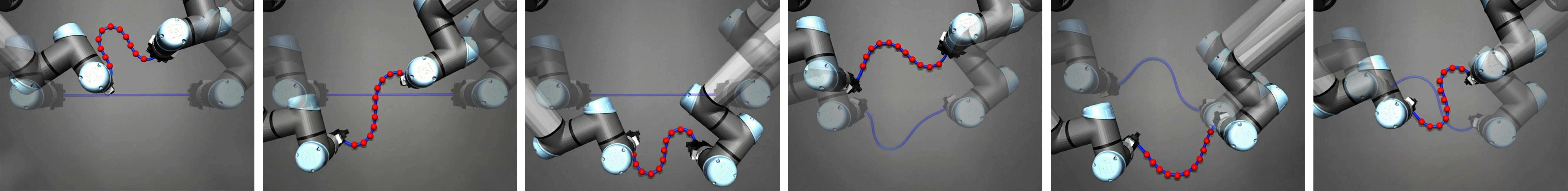}
    \end{subfigure}
  \end{minipage}
  
  
  \begin{minipage}[c]{\textwidth}
    \centering
    \makebox[0pt][r]{\raisebox{0.5\height}{(b)}\hspace{6pt}}%
    \begin{subfigure}[c]{0.92\textwidth}
      \centering
      \includegraphics[width=\linewidth]{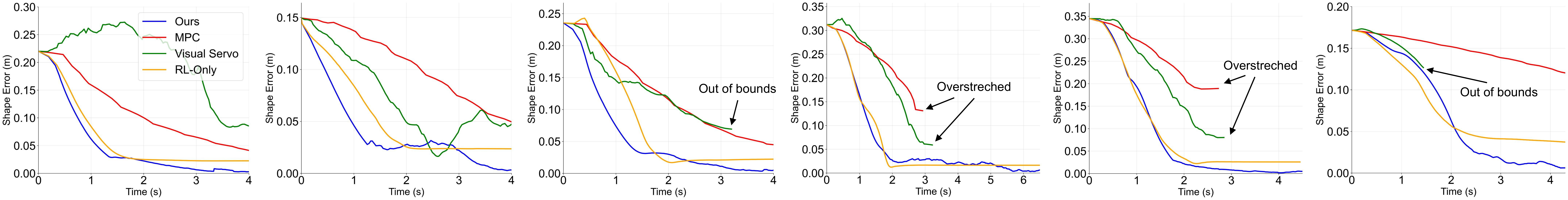}
    \end{subfigure}
  \end{minipage}
  
  \caption{Representative shape control cases in simulation. (a) Six test cases showing the initial (transparent) and final (opaque) DLO shapes. Red dots denote the target shape. (b) Corresponding shape error curves of different methods. (Refer to our supplementary video for the full control processes.)}  
  \label{fig:fig7}
\end{figure*}

Our method achieves the highest success rate and convergence speed under both initial conditions. Ablating the difficulty filtering module leads to poor training signals and reduced sample efficiency, causing noticeable instability and performance collapse under the straight-line condition, which validates the necessity of a structured easy-to-hard curriculum. The variant without diversity selection also underperforms, demonstrating that goals with high spatial coverage and information richness promote better policy generalization. Complete removal of curriculum learning yields poor performance, further confirming the necessity of structured curriculum guidance. In contrast, our approach jointly enforces difficulty and diversity, continuously supplying reachable yet challenging goals, which yields more stable training and superior final performance.

Moreover, the model-free baseline performs significantly worse in both conditions with slow learning progress, highlighting its inherent sample inefficiency, while our model-based approach substantially improves training efficiency through the use of model-generated synthetic data.

Fig.~\ref{fig:fig6} illustrates the evolution of the self-curriculum goal generation. Initially (Epoch~0), the generated goals cluster around the initial shape. As training advances (Epochs~20, 100), the goals become more complex and spread more uniformly in the workspace. The right plot confirms a steady growth in the average Euclidean distance between goals and the initial shape across epochs. These results show that the self-curriculum mechanism automatically produces goals of increasing difficulty and diversity, providing well-matched challenges as the policy improves.

\subsection{Shape Control Strategies}

We further compare our two-stage shape control strategy with three representative baselines:
(1) Gradient-based MPC ~\cite{gu2025learning,yang2022online,li2018learning}: it uses one sub-model from our learned ensemble as the forward dynamics predictor and optimizes an $H$-step control sequence by SGD-based minimization of cumulative shape error.
(2) Jacobian-based visual servo ~\cite{zhu2018dual,jin2019robust}: it is the same controller as our small-deformation stage, but is applied from the initial shape.
(3) RL-only: it follows the same RL training framework as our large-deformation stage, but is retrained under a stricter success threshold and executed without the visual servo refinement.

For our method, the stage switching threshold is set to $\epsilon=30~\mathrm{mm}$ and the visual servo gain $\lambda$ is set to $0.05$. For MPC, we use a planning horizon of $H=5$. For RL-only, we set the training success threshold to $\epsilon_{\mathrm{RL}}=3~\mathrm{mm}$. These parameters were selected through preliminary trials and then fixed across all test cases. We evaluate all methods on the same test sets introduced in Sec.~B under both initial conditions. To assess generalization, we test with the DLO's bending stiffness increased by 10 times. 

During evaluation, each method is allowed to execute up to 250 steps. We declare task success if the shape error $e$ falls below $10~\mathrm{mm}$. We additionally report three metrics: (i) the average minimum shape error achieved over the entire trajectory for all test cases, (ii) the average minimum shape error over successful cases, and (iii) the average time required for successful cases to reach the success threshold.

As summarized in Table~\ref{tab:sim_comparison}, our method achieves the highest success rate and the lowest average minimum shape error under both initial conditions, significantly outperforming all baseline methods.

\begin{table}[t]
\centering
\caption{Simulation comparison of shape control strategies under straight and diverse initializations.}
\label{tab:sim_comparison}
\renewcommand{\arraystretch}{1.15}
\setlength{\tabcolsep}{3.5pt}

\begin{tabular}{c c c c c c}
\toprule
Scenario &
Method &
\makecell{Avg. min.\\ error (all)\\ (mm)$\downarrow$} &
\makecell{Success\\ rate$\uparrow$} &
\makecell{Avg. min.\\ error (succ.)\\ (mm)$\downarrow$} &
\makecell{Avg. time\\ (succ.)\\ (s)$\downarrow$} \\
\midrule
\multirow{4}{*}{\makecell{Straight\\ Initial}}
& Ours         & \textbf{4.678} & \textbf{47/50} & \textbf{1.696} & 2.147 \\
& MPC          & 43.15          & 30/50          & 2.342          & 6.592 \\
& Visual Servo & 47.37          & 13/50          & 2.862          & 1.959 \\
& RL-Only           & 29.69          & 10/50          & 7.835          & \textbf{1.450} \\
\midrule
\multirow{4}{*}{\makecell{Diverse\\ Initial}}
& Ours         & \textbf{12.49} & \textbf{43/50} & \textbf{2.204} & 2.464 \\
& MPC          & 36.45          & 26/50          & 2.395          & 4.907 \\
& Visual Servo & 44.06          & 18/50          & 3.295          & 2.783 \\
& RL-Only           & 27.82          & 10/50          & 8.276          & \textbf{1.925} \\
\bottomrule
\end{tabular}
\end{table}

MPC performs well on some simple cases, but its performance strongly depends on the accuracy of the dynamics model. In particular, when the DLO is close to a straight configuration, model prediction errors can induce overstretched motions and cause failures. Furthermore, its online gradient-based optimization is computationally expensive, leading to long execution times. The Visual Servo baseline relies heavily on Jacobian initialization; under large deformations, it is prone to local minima and often exhibits unstable motions, leading to low success rates. RL-Only attains a lower average minimum error than MPC and Visual Servo, showing potential for large deformations. However, it achieves the lowest success rate and the worst accuracy among its successful cases, suggesting difficulty in achieving precise convergence. In contrast, our method only requires learning how to move the DLO roughly towards the target, without requiring extreme precision, which significantly simplifies the training process.

Fig.~\ref{fig:fig7} presents concrete task examples that further support this analysis. Our method efficiently and precisely accomplishes various large-deformation tasks, including those with opposite curvatures, exhibiting smooth and rapidly converging error curves. In contrast, the baselines frequently show oscillating error profiles, premature termination (indicating task failure), or convergence to suboptimal, high-error states.

\section{REAL-WORLD EXPERIMENT AND RESULTS}

\subsection{Experiment Setup}

\begin{figure}[t]
    \centering
    \includegraphics[width=0.90\columnwidth]{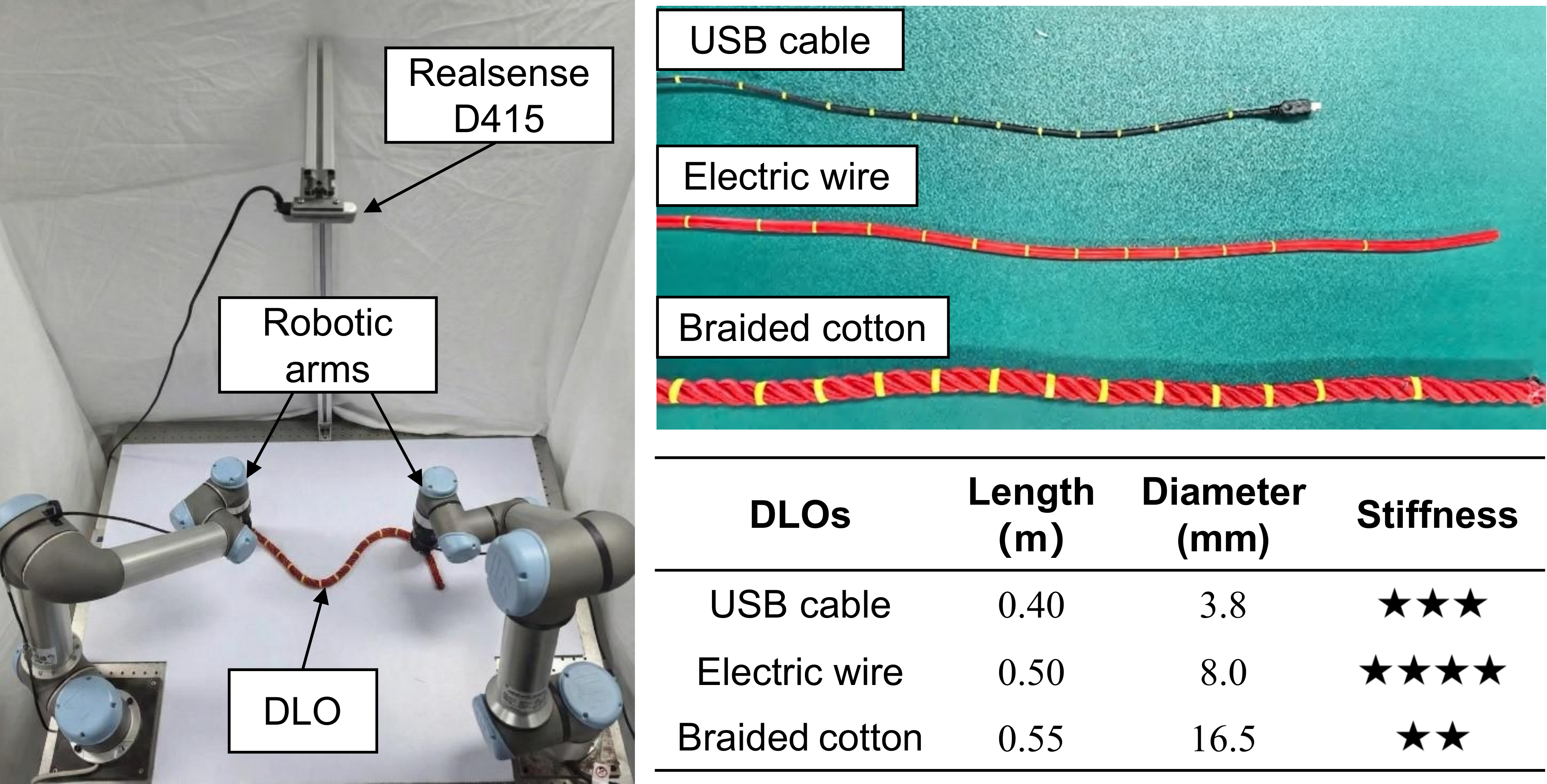}  
    \caption{Real-world experiment setup and DLOs used in the experiment.}
    \label{fig:fig8}
\end{figure}

The real-world experimental setup is illustrated in Fig.~\ref{fig:fig8}. A UR5 arm (left) and a UR5e arm (right) are used to grasp the two ends of the DLO. Consistent with the simulation, the DLO shape is represented by 13 markers, including the two gripping points. An Intel RealSense D415 camera, mounted overhead, tracks the keypoints in real time via color segmentation. To validate the method, three DLOs with different sizes and material properties are tested: an electric wire (purely elastic), a USB cable, and a braided cotton rope, the latter two exhibiting certain plasticity. Real-world experiments are conducted on a PC with an Intel Core i5-7400 CPU and an NVIDIA GeForce GTX 1050Ti GPU.

\subsection{Shape Control Strategies on Various DLOs}

We evaluate the proposed method on each DLO with a total of 10 task cases, whose target shapes are pre-collected and guaranteed to be achievable~\cite{yu2022global}. These cases comprise two initial conditions (5 cases each). Under the diverse initial condition, half of the cases involve opposite‑curvature deformations. For the large-deformation stage, we directly used the policy trained in simulation, without collecting any real DLO data for retraining or fine-tuning.

Comparison results are shown in Table~\ref{tab:real_comparison}. Our method successfully completed all 30 test cases with high accuracy across three different DLOs, demonstrating strong sim2real transfer capability. This robustness stems from our two-stage design: the RL policy only needs to  provide coarse, generalizable directional guidance learned in simulation, while the model-free visual servoing controller naturally generalizes to real-world scenarios, effectively avoiding sim2real errors.

\begin{table}[t]
\centering
\caption{Real-world comparison of shape control strategies under straight and diverse initializations.}
\label{tab:real_comparison}
\renewcommand{\arraystretch}{1.15}
\setlength{\tabcolsep}{3.5pt}

\begin{tabular}{c c c c c c}
\toprule
Scenario &
Method &
\makecell{Avg. min.\\ error (all)\\ (mm)$\downarrow$} &
\makecell{Success\\ rate$\uparrow$} &
\makecell{Avg. min.\\ error (succ.)\\ (mm)$\downarrow$} &
\makecell{Avg. time\\ (succ.)\\ (s)$\downarrow$} \\
\midrule
\multirow{4}{*}{\makecell{Straight\\ Initial}}
& Ours         & \textbf{1.985} & \textbf{15/15} & 1.985 & \textbf{14.285} \\
& MPC          & 12.749         & 12/15          & \textbf{1.780}          & 24.880 \\
& Visual Servo       & 38.648         & 4/15           & 5.759          & 17.168 \\
& RL-Only      & 15.719         & 6/15           & 7.130          & 14.380 \\
\midrule
\multirow{4}{*}{\makecell{Diverse\\ Initial}}
& Ours         & \textbf{2.064} & \textbf{15/15} & \textbf{2.064} & 20.434 \\
& MPC          & 23.238         & 8/15           & 3.070          & 40.399 \\
& Visual Servo       & 60.745         & 2/15           & 4.702          & 33.426 \\
& RL-Only      & 38.313         & 3/15           & 8.535          & \textbf{10.147} \\
\bottomrule
\end{tabular}
\end{table}

\begin{figure*}[t!]   
  \centering
  \begin{minipage}[c]{\textwidth}
    \centering
    \makebox[0pt][r]{\raisebox{0.5\height}{(a)}\hspace{6pt}}%
    \begin{subfigure}[c]{0.88\textwidth}
      \centering
      \includegraphics[width=\linewidth]{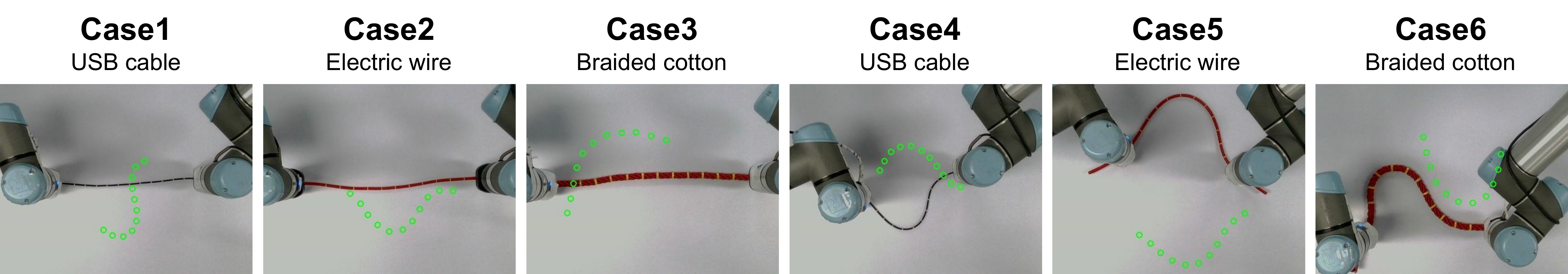}
    \end{subfigure}
  \end{minipage}
  
  
  \begin{minipage}[c]{\textwidth}
    \centering
    \makebox[0pt][r]{\raisebox{0.5\height}{(b)}\hspace{6pt}}%
    \begin{subfigure}[c]{0.88\textwidth}
      \centering
      \includegraphics[width=\linewidth]{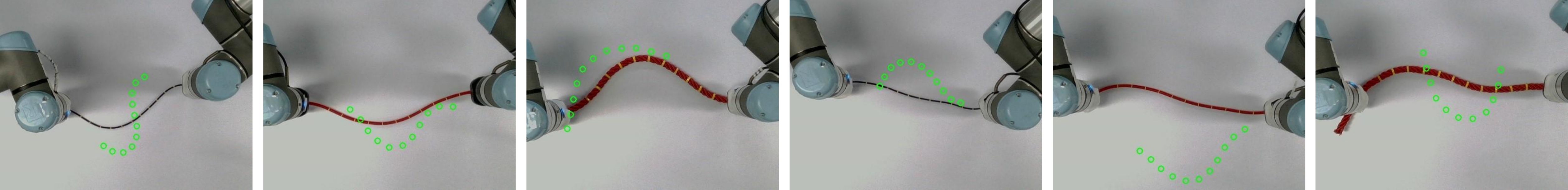}
    \end{subfigure}
  \end{minipage}
  
  \begin{minipage}[c]{\textwidth}
    \centering
    \makebox[0pt][r]{\raisebox{0.5\height}{(c)}\hspace{6pt}}%
    \begin{subfigure}[c]{0.88\textwidth}
      \centering
      \includegraphics[width=\linewidth]{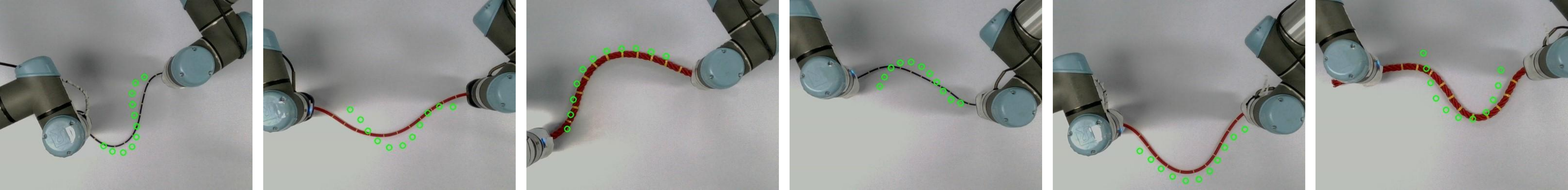}
    \end{subfigure}
  \end{minipage}

  \begin{minipage}[c]{\textwidth}
    \centering
    \makebox[0pt][r]{\raisebox{0.5\height}{(d)}\hspace{6pt}}%
    \begin{subfigure}[c]{0.88\textwidth}
      \centering
      \includegraphics[width=\linewidth]{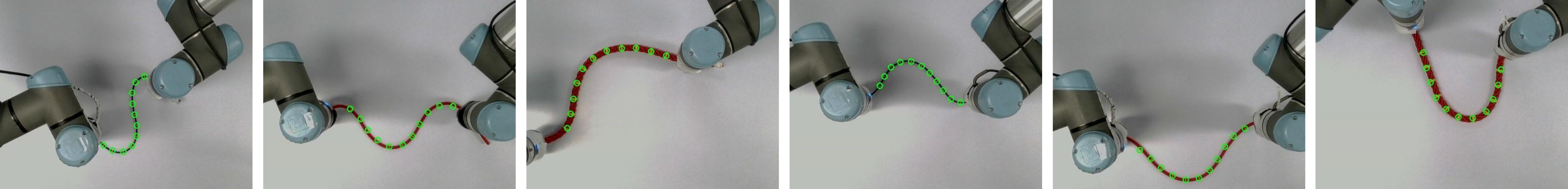}
    \end{subfigure}
  \end{minipage}

  \begin{minipage}[c]{\textwidth}
    \centering
    \makebox[0pt][r]{\raisebox{0.5\height}{(e)}\hspace{6pt}}%
    \begin{subfigure}[c]{0.88\textwidth}
      \centering
      \includegraphics[width=\linewidth]{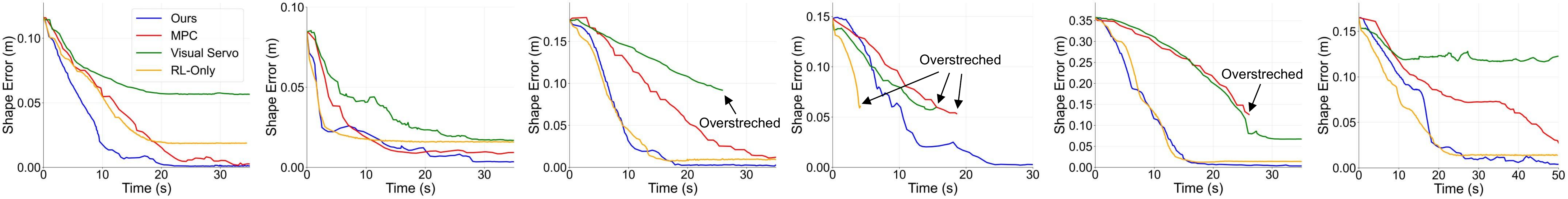}
    \end{subfigure}
  \end{minipage}
  
  \caption{Real-world shape control process of six cases. (a) Initial shape. (b) Shape at the large deformation stage. (c) Transition point between large and small deformation stages. (d) Final shape achieved. (e) Corresponding shape error curves of different methods. Green circles represent the target shape. (Refer to our supplementary video for the full control processes.)}  
  \label{fig:fig9}
\end{figure*}

In contrast, the MPC method suffered from sim2real errors, leading to suboptimal performance, particularly when handling cases with opposite curvature. Additionally, in environments with more limited computational resources, the execution efficiency of MPC becomes more pronounced. The Visual Servo method, although model‑free and generalizable, still performs poorly under large deformations. Similar to the simulation, the RL-Only method failed to achieve high-precision control; it also struggled with some opposite-curvature tasks in these real-world trials, indicating that training with a low success threshold yields suboptimal policies due to sparse reward signals. The real-world shape control process and error curves are shown in Fig.~\ref{fig:fig9}.

\section{CONCLUSIONS}

This paper presents a two-stage framework combining MBRL and visual servoing for precise DLO shape control under large deformations. Its effectiveness stems from a clear task decomposition: the RL policy handles the long-horizon challenge of large deformations, while the model-free visual servoing controller ensures precise, real-world convergence. Extensive experiments validate the method's high success rate, precision, and strong sim-to-real generalization across diverse DLOs. The self-curriculum goal generation mechanism enables this performance by facilitating sample-efficient learning of a robust policy. Moreover, this self-curriculum paradigm, which dynamically generates goals through imagined evaluation, offers a promising strategy for other multi-goal scenarios characterized by varied initial states.

While the current framework is validated in 2D settings, future work will explore its extension to 3D environments, which presents challenges such as increased action space complexity and higher demands on modeling accuracy.

\addtolength{\textheight}{0cm}   





\bibliographystyle{IEEEtran}  
\bibliography{IEEEabrv, references}     

\end{document}